\documentclass[final,onefignum,onetabnum]{siamart171218}

\usepackage{lipsum}
\usepackage{amsfonts}
\usepackage{graphicx}
\usepackage{epstopdf}
\usepackage{amsmath}
\usepackage{amssymb}
\usepackage{calligra}
\usepackage{calrsfs}
\usepackage{algorithm,algorithmic}
\usepackage{booktabs}
\usepackage{subfig}
\usepackage{epsfig}
\usepackage{color}

\def\divg{\rm div}

\ifpdf
  \DeclareGraphicsExtensions{.eps,.pdf,.png,.jpg}
\else
  \DeclareGraphicsExtensions{.eps}
\fi

\newsiamremark{remark}{Remark}
\newsiamremark{hypothesis}{Hypothesis}
\crefname{hypothesis}{Hypothesis}{Hypotheses}
\newsiamthm{claim}{Claim}

\headers{Convex hull algorithm}{L. Li, S. Luo, XC. Tai and J. Yang}

\title{CONVEX HULL ALGORITHMS BASED ON SOME VARIATIONAL MODELS\thanks{Submitted to the editors DATE.}}

\author{Lingfeng Li\thanks{Department of Mathematics, Hong Kong Baptist University, Hong Kong, China; Department of Mathematics, Southern University of Science and Technology, Shenzhen, China}
\and Shousheng Luo\thanks{Beijing Computational Science Research Center, Beijing, China;
School of Mathematics and Statistics, Data Analysis Technology Lab, Henan University, Kaifeng, China}
\and Xue-Cheng Tai\thanks{Department of Mathematics, Hong Kong Baptist University, Hong Kong, China (\email{xuechengtai@hkbu.edu.hk})} 
\and Jiang Yang\thanks{Department of Mathematics, Southern University of Science and Technology, Shenzhen, China}
}

\usepackage{amsopn}

\makeatletter
\newcommand*{\addFileDependency}[1]{
  \typeout{(#1)}
  \@addtofilelist{#1}
  \IfFileExists{#1}{}{\typeout{No file #1.}}
}
\makeatother

\newcommand*{\myexternaldocument}[1]{%
    \externaldocument{#1}%
    \addFileDependency{#1.tex}%
    \addFileDependency{#1.aux}%
}

\usepackage{color}

\ifpdf
\hypersetup{
  pdftitle={CONVEX HULL ALGORITHMS BASED ON SOME VARIATIONAL MODELS},
  pdfauthor={L. Li, S. Luo, XC. Tai and J. Yang}
}
\fi

\myexternaldocument{ex_supplement}

\begin{document}

\maketitle

\begin{abstract}
Seeking the convex hull of an object is a very fundamental problem arising from various tasks. In this work, we propose two variational convex hull models using level set representation for 2-dimensional data. The first one is an exact model, which can get the convex hull of one or multiple objects. In this model, the convex hull is characterized by the zero sublevel-set of a convex level set function, which is non-positive at every given point. By minimizing the area of the zero sublevel-set, we can find the desired convex hull. The second one is intended to get convex hull  of objects with outliers. Instead of requiring all the given points are included, this model penalizes the distance from each given point to the zero sublevel-set. Literature methods are not able to handle outliers. For the solution of these models, we develop  efficient numerical schemes using alternating direction method of multipliers. Numerical examples are given to demonstrate the advantages of the proposed methods.
\end{abstract}



\section{Introduction}
\begin{figure}[ht]
    \centering
    \subfloat[A given points set]{\includegraphics[width=4cm]{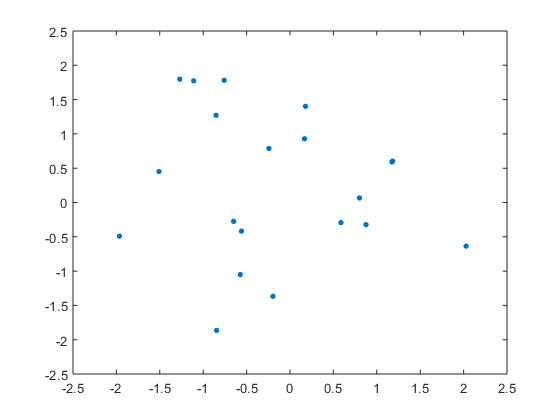}}
    \subfloat[Convex hull of the set]{\includegraphics[width=4cm]{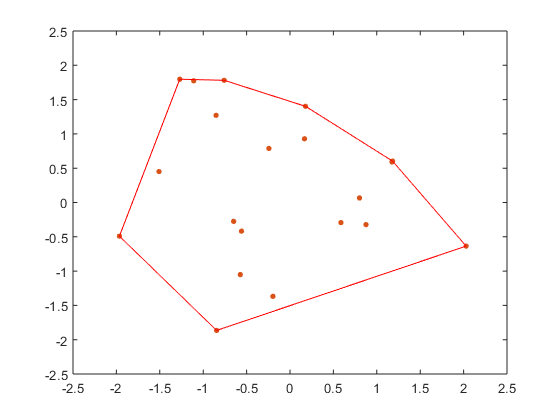}}
    \caption{An example of convex hull.}
    \label{fig:convex_hull_example}
\end{figure}

Seeking convex hull is one of fundamental problems in computational geometry. 
The convex hull of a set $\mathbb{S}$ is defined as the smallest convex set containing $\mathbb{S}$. In other words, every convex set containing $\mathbb{S}$ also contains the convex hull. An example is shown in Figure \ref{fig:convex_hull_example}.

Convex hull problems arise from different areas, such as data clustering \cite{LiparuloLuca2015FCUt}, robot motion planning \cite{HertS.1999MpiR}, collision detection \cite{TomicTeodor2017EWEC}, image segmentation \cite{Condat2017}, disease diagnosis \cite{zhang2009convex} and  robust estimation \cite{LiuReginaY.1999Mabd}. When the input data is in a 2-dimensional space, which is the most common case, the convex hull of a finite set is a polygon. When the input data is in an n-dimensional ($n>2$) space, the convex hull of a finite set is an n-polytope. Usually, convex hull algorithms take the coordinates of points as input, and yield the vertices of the associated convex hull polygon. 

Various convex hull algorithms were proposed in the literature. The earliest convex hull algorithm, as far as we know, is the gift wrapping method for 2D input proposed by Chand and Kapur in \cite{chand1970algorithm} and Jarvis in \cite{jarvis1973} independently.  This method first determines the leftmost point, which is a vertex of the convex hull, and then search for the point that every other point lies on one side of the line from the current vertex to it. Obviously, the point found in this way must be one vertex of the associated convex hull. Repeating this procedure, one can find all the vertices of the convex hull. 
A modified version is the Gram scan \cite{graham1972} proposed by Gram. Starting from the lowest point, all the points are sorted in increasing order of the angle they and the lowest point make with the x-axis, and then a more efficient searching scheme can be performed. 
A variant of the Gram scan is the monotone chain algorithm proposed by Andrew \cite{andrew1979}. Instead of sorting points by the angles, this algorithm sorts points by their coordinates. In \cite{BarberC1996Tqaf}, the quickhull method was developed by Barber, Dobkin and Huhdanpaa. Similar to the quicksort algorithm \cite{hoare1961algorithm}, the quickhull algorithm divides the problem into many subproblems recursively and solves them independently. Another algorithm using the same idea is the 
"divide and conquer" algorithm by Preparata and Hong \cite{PreparataF1977Chof}. There are also other convex hull algorithms, such as the incremental convex hull algorithm by Kallay \cite{kallay1984complexity}, the ultimate planar convex hull algorithm by Kirkpatrick and Seidel \cite{kirkpatrick1986ultimate} and Chan's algorithm \cite{chan1996optimal}.

In terms of the computational complexity, the gift wrapping method \cite{chand1970algorithm,jarvis1973} takes $O(nh)$, where $n$ is the number of points given, and $h$ is the number of vertices of the associated convex hull polygon. The Gram scan \cite{graham1972}, monotone chain \cite{andrew1979}, quick hull \cite{BarberC1996Tqaf}, divide and conquer \cite{PreparataF1977Chof}, and the incremental algorithm \cite{kallay1984complexity} all have the same complexity  $O(n\log{n})$. If the given data is already sorted in the desired order, the monotone chain and Gram scan take only $O(n)$. The ultimate planar \cite{kirkpatrick1986ultimate} and Chan's algorithm \cite{chan1996optimal} both take $O(n\log{h})$.

For some applications, one may only need an approximate convex hull instead of an exact one. For example, seeking all the convex hull vertices may be relatively expensive if the exact convex hull contains too many vertices. To address this problem, many approximating algorithms were developed. Commonly, only a subset rather than the whole set is used to compute the convex hull. In \cite{BentleyJon1982Aafc}, Bentley, Preparata and Faust proposed a planar algorithm. This method first divides the input data into many strips based on their x-axis coordinates and takes the lowest and highest points in each strip to form a subset. Then, various exact convex hull algorithms can be conducted on this new subset. A similar algorithm was given by Kavan, Kolingerova and Zara in \cite{kavan2006fast}, where they separate the points into different sections based on the angles they and the origin make with the x-axis, and the out-most point in each section is picked for computing the convex hull. Similar methods can also be found in \cite{klette1983approximation}, \cite{vzunic1990approximate} and \cite{krvr1995sequential}. Another new model based on the active contour was proposed by Sirakov in \cite{sirakov2006new}. This method uses a parameterized curve to approximate the boundary of convex hulls and prevents the concavity by introducing a special vector field. 

A special type of input data for the convex hull problem is binary masks, where a point with value 0 represents the point to be enclosed. In image processing, binary masks are often obtained by some image segmentation techniques. If we extract the coordinates of all the pixels with value 0, we can also apply the algorithms mentioned above to find the convex hull. However, instead of just identifying the convex hull vertices, we may also need to draw the whole region of the convex hull. When the algorithms  only return the vertices, we need extra steps to determine the convex hull polygon region. This can be done by some fast algorithms, for example, the scan line algorithm \cite{hearn2010computer}. 

\begin{figure}[ht]
    \centering
    \subfloat[True convex hull]{\includegraphics[width=3.5cm]{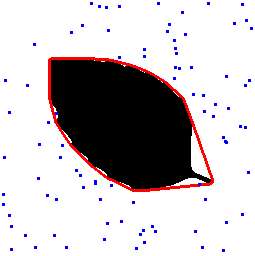}}\quad
    \subfloat[Result by quickhull]{\includegraphics[width=3.5cm]{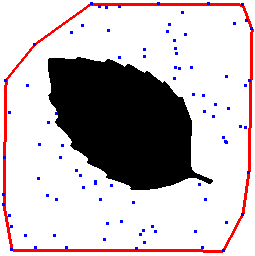}}\quad
    \subfloat[Result by the proposed method]{\includegraphics[width=3.5cm]{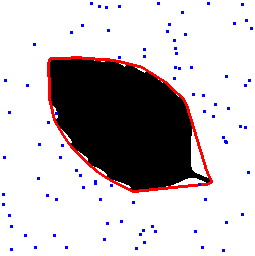}}
    \caption{Convex hulls of objects polluted by noise.}
    \label{fig:noisy_hull_example}
\end{figure}

Another important issue is the existence of noise and outliers in the given points, which is often indispensable in real applications. For example, if the image is noisy or the background is complicated, the binary masks obtained may contain outliers, which will dramatically change the shape and size of the convex hull. Figure \ref{fig:noisy_hull_example} illustrates a simple example of how the outliers affect the convex hull. Here the binary image is about a leaf polluted by noise. 
In real applications, the convex hull of leaves can be used for species recognition \cite{du2007leaf}. 

However, all the existing methods are not able to handle the situation with noise and outliers. One way to deal with this problem is mentioned in \cite{biro2012approximation}. Briefly speaking, we need to check all the subsets of the original set with $N-k$ elements and find the one that has the smallest convex hull area, where $N$ is the total number of points given and $k$ is the number of outliers we assume. Theoretically, this method can filter out all the outliers exactly if we know the number of outliers, but this is usually not the case. What's more, the computational cost is very high when $k$ is large.

In this paper, we propose two variational convex hull models based on the level set representation for 2-dimensional data. Extension to high dimensional data is easy.  Level set method is a widely studied tool in image processing \cite{chan1999active}\cite{tai2017simple}, because it is able to track the change of topology. It is well known that any region can be characterized by the zero sublevel-set of its signed distance function (SDF), which is a special level set function. Therefore, finding the convex hull of a region is equivalent to seeking its corresponding SDF. The first model we proposed is for exact convex hulls. In this case, we obtain the exact convex hulls by minimizing the area of the zero sublevel-set of a convex SDF. The SDF is also required to be non-positive on the zero sub-level set region. To impose the convexity constraint, we require the Laplacian of the SDF is non-negative at the given set. The equivalence of these two conditions was proved in \cite{yan2018convexity} and \cite{luo2018convex}. The second model we introduced is for the cases with outliers or noise, where the convex hulls do not have to enclose all the given points. Instead, we penalize the distance from each given point to the zero sublevel-set in the objective function. What's more, both models can be modified to find the convex hulls of multiple objects simultaneously.

An efficient numerical method is developed to solve the proposed models. Without changing the optimal solution, we modify the objective function, area of zero sublevel-set, to another form which is easier to solve. Then alternating direction method of multipliers (ADMM) is applied to solve the constrained optimization problems for exact and inexact models. For computational efficiency, we further assume that the data is periodic in the 2D spatial coordinate which infers 
the associated solution is periodic SDF functions \cite{luo2018convex}. 
Due to this, the partial differential equations arising from the ADMM method can be solved by applying fast Fourier transform (FFT).  One result of our convex hull algorithm is shown in Figure \ref{fig:noisy_hull_example} (c).

The rest part of this paper is organized as follows. In Section \ref{sec2}, we will give a brief introduction to convex hull and level set method. In Section \ref{sec3}, we will develop our convex hull models in detail and explain the rationale behind. In Section \ref{sec4}, an efficient numerical algorithm will be provided. In Section \ref{sec5}, we will conduct some numerical experiments for our models.

\section{Preliminaries}\label{sec2}
Suppose the object given is inside a rectangular domain $\Omega=[0,M]\times[0,N]\subset \mathbb{R}^2$, and the binary function $I(x):\Omega\rightarrow\{0,1\}$ is the indicator function of an open subset, denoted by $\Omega_0\subset\Omega$. Then, we want to find its convex hull Conv$(\Omega_0)$, or equivalently the indicator function of Conv$(\Omega_0)$. From the later formulations, one will see that there is no problem for our approach if the subset $\Omega_0$ only contains some isolated points. 
To represent a region in $\Omega$, one efficient way is using the level-set representation method. A level-set function is defined on the whole domain $\Omega$ and takes 
values in $\mathbb{R}$: $\phi(x):\Omega\rightarrow \mathbb{R}$. Given a level-set function, its sublevel-set $\text{slev}_{\phi}^c=\{x|\phi(x)< c\}$ is a subset of $\Omega$. Usually, we use the zero sublevel-set $\text{slev}_{\phi}^0$ to characterize the region of interest. Therefore, instead of seeking the convex hull, we can search for its corresponding level-set function $\phi(x)$, and then recover it by taking the zero sublevel-set of $\phi(x)$. 

\begin{figure}
    \centering
    \subfloat[]{\includegraphics[width=4cm]{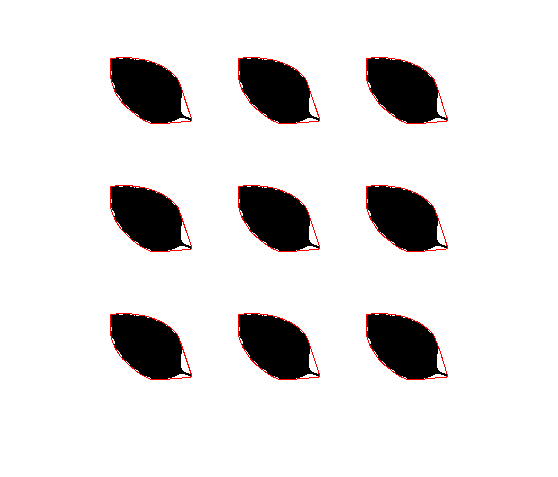}}
    \subfloat[]{\includegraphics[width=4cm]{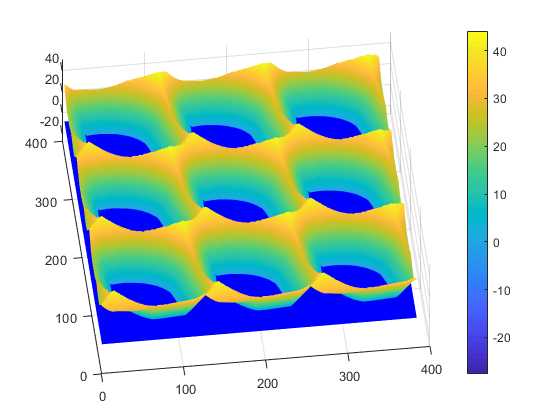}}
    \subfloat[]{\includegraphics[width=4cm]{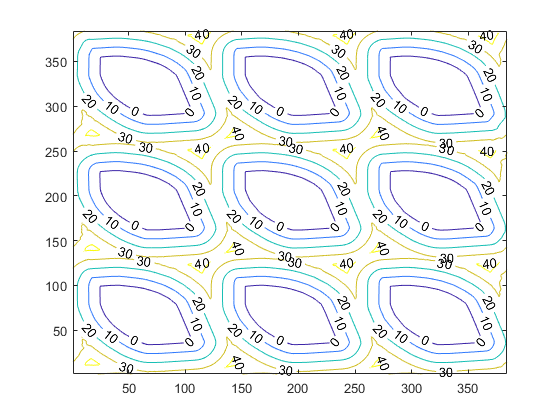}}
    \caption{(a) shows periodic replications of a leaf image and their convex hulls, (b) shows the computed $\phi$ of the convex hulls, and (c) plots the level-set curves of $\phi$.}
    \label{fig:periodic_assuamption}
\end{figure}

One popular choice of level-set function is the signed distance function (SDF). Given a subset $\Omega_0$ with piecewise smooth boundary, its corresponding SDF is defined as
\begin{equation}
    \phi(x)=\begin{cases}
    -\text{dist}(x,\partial \Omega_0),\quad &x\in\Omega_0\\
    \text{dist}(x,\partial \Omega_0),\quad &x\notin\Omega_0
    \end{cases}.\label{eq:SDF}
\end{equation}
Notice that $\phi(x)=0$ when $x$ is exactly on $\partial\Omega_0$, which represents the boundary of $\Omega_0$. To simplify the notation, we will use $\phi(x)$ to denote a SDF in the following text. One important property of SDF is that
\begin{equation}
    |\nabla\phi(x)|=1\ a.e. \mbox{  in  } \Omega.\label{eq:norm_gradient}
\end{equation}
One can see that different open subsets are corresponding to different SDFs. In this work, we further assume the data $I(x)$ is periodic in $\mathbb{R}^2$. Correspondingly, all other functions that we will need to compute are also periodic in $\mathbb{R}^2$. When they are discretized and computed in $\mathbb{R}^2$, they will be periodic in both $x_1$ and $x_2$ directions. If we denote the union of all periodic replications of the objects as $\tilde{\Omega}$, in this case, the distance in (\ref{eq:SDF}) is the distance to $\partial\tilde{\Omega}$. To make it easier to understand this assumption, we use the leaf example to illustrate it in Figure \ref{fig:periodic_assuamption}. 

For the convex hull problem, what we want to do is to seek the smallest convex set containing $\Omega_0$. Based on the discussion above, we can search for the SDF of the convex hull instead of the convex hull directly, and recover the convex hull region from the SDF. Because the convex hull is a convex set, we should add this constraint into the model. In a 2-dimensional space, the relation between the convexity of a subset and its corresponding SDF is characterized by the following theorem \cite{luo2018convex}:
\begin{theorem}\label{th:convex_cond}
Let $\phi$ be the SDF of a subset $\Omega_1\subset\Omega$ and $\phi\in C^2(\Omega)$ almost everywhere. If $\Omega_1$ is convex, then $\phi$ must satisfy:
\begin{align}
\Delta\phi\geq 0 \text{ a.e. in }  \Omega.
\end{align} 
\end{theorem}
This can be easily seen by computing the curvature of the level-set curves, which is defined as the boundary of $\text{slev}^c_{\phi}=\{x|\phi(x)\le c\}$. If we further assume $|\nabla\phi|\neq 0$ a.e. in $\Omega$, then the curvature of the level-set curves passing through $x$ can be computed as
\begin{equation}
    k=\divg\left(\frac{\nabla\phi}{|\nabla\phi|}\right)=\Delta\phi.
\end{equation}
The equality holds because of the property (\ref{eq:norm_gradient}). If $\Omega_1$ is convex, then it is well known that the curvature $k$ is non-negative at points the SDF is smooth. In case the SDF is not smooth at some points, we can define  $\Delta \phi$ in the weak sense and it is still true that $\Delta \phi \ge 0$ in $\Omega$ if $\Omega_1$ is convex. In this work, since we assume $\phi(x)$ is periodic, it is not possible to have $\Delta\phi\geq 0$ everywhere. Instead, we will ask $\Delta\phi\geq 0$  only  in a subregion of $\Omega$, i.e, 
\begin{equation}
    \Delta\phi\geq 0 \text{ a.e. in } \text{slev}^c_{\phi} \text{ for some } c\geq 0.\label{eq:convex_condition}
\end{equation}
For example, in Figure \ref{fig:periodic_assuamption}, the level-set curves are convex only for $c$ up to 20. The idea to require the level set function to satisfy $\Delta\phi\geq 0$ only in a subregion to guarantee the convexity of the represented region  is also used in \cite{luo2019} for representation of multiple convex objects. It is enough to choose $c=0$ to guarantee the represented region is convex. Later, we shall see that larger values of $c$ will also help to merge separated convex regions which are close to each other.

In order to obtain the associated SDF of the convex hull of given points, we should require the SDF to be non-positive on $\Omega_0$, i.e., 
\begin{equation}
     \phi(x)\leq 0 \text{ for } \forall x\in \Omega_0. \label{eq:enclose_condition}
\end{equation}
Then, with the condition (\ref{eq:norm_gradient}), (\ref{eq:convex_condition}) and (\ref{eq:enclose_condition}), we can characterize the convex hull problem mathematically in the next section.

\section{Model description}\label{sec3}
\subsection{Exact convex hull model}
Let's first consider the case where only one object is contained in the given binary image. From the definition of the convex hull, we know that the convex hull of a subset $\Omega_0$ is the smallest convex set containing $\Omega_0$, so $\text{Conv}(\Omega_0)$ can be viewed as the minimizer of the following optimization problem:
\begin{align}
    \underset{\Omega_1\subseteq \Omega}{\min}\quad &\text{Area}(\Omega_1) \label{eq:convhull1}\\
    {\rm subject\ to}\quad& \Omega_1 \text{ is convex,} \label{eq:constraint_1}\\
    & \Omega_0\subseteq \Omega_1. \label{eq:constraint_2}
\end{align}
According to the previous discussion, we can solve the convex hull problem by searching for its corresponding SDF. Suppose $\phi$ is the SDF of a set $\Omega_1$, then its area can be written as $\int_{\Omega}[1-H(\phi(x))]dx$, where $H(s)$ is the Heaviside function:
\begin{equation}
    H(s)=\begin{cases}
    1 & s\geq 0\\
    0 & s<0.
    \end{cases}
\end{equation}

In fact, there are two equivalent functionals we can use for this problem. The first one is $\int_{\Omega}-\phi(x)dx$. 
Suppose there is another convex set containing $\Omega_0$ with SDF $\phi_1$. Because the convex hull is the smallest convex set containing $\Omega_1$, we can easily show that $\phi(x)\geq \phi_1(x)$ at every point in $\Omega$. Consequently, the minimizer of $\int_{\Omega}-\phi(x)dx$ with the same constraints is also the SDF of the convex hull. The second one is the boundary length of $\Omega_0$, which can be represented as $\int_{\Omega}|\nabla H(\phi)|dx$, since the convex hull has the shortest boundary length among the convex sets containing $\Omega_0$. A sketch of the proof can be found in \cite{berg2008computational}. Based on our numerical test, we choose
\begin{equation}
    \int_{\Omega}-\omega\phi(x)+\mu|\nabla H(\phi(x))|
\end{equation}
as the objective functional, where $\omega, \mu>0$ are two user-specified parameters. The reason we don't use $H(\phi)$ as the objective functional is that its derivative with respect to $\phi$ will disappear except on the zero level-set curve, which will result in low convergence rate. Using $-\phi(x)$ can make the algorithm converge much faster. The total variation term here can help regularize the boundary of convex hulls in some special cases. Furthermore, if we have some knowledge about the position of the convex hull boundary, e.g., the leftmost point must lie on the boundary of the convex hull, we can incorporate this information into our model by adding $\sum_{\hat{x}_i}\phi(\hat{x}_i)^2$ to the model, where $\hat{x}_i$ are some assigned boundary landmarks. Based on the previous discussion, we formulate the exact convex hull model as
\begin{align}
    \underset{\phi}{\min}\quad &\int_{\Omega}F_1(x,\phi)dx\label{eq:convhull2}\tag{EC}\\
    {\rm subject\ to}\quad&|\nabla\phi(x)|=1 \text{ a.e. in }\Omega\label{eq:constraint_3}\\
    & \Delta\phi(x)\geq 0 \text{ a.e. in } \text{slev}_{\phi}^c\label{eq:constraint_4}\\
    & \phi(x)\leq 0 \text{ for every }x\in \Omega_0, \label{eq:constraint_5}
\end{align}
where $F_1(x,\phi)=-\omega\phi(x)+\mu|\nabla H(\phi(x))|+\nu \sum_{\hat{x}_i}\phi(\hat{x}_i)^2$ with $\nu>0$. In order to accelerate the convergence of our algorithm, we can increase the value of $c$ in (\ref{eq:constraint_4}). However, this $c$ can not be larger than the margin of the image, i.e, the minimum distance between the object $\Omega_0$ and the domain boundary $\partial\Omega$. Otherwise, the algorithm may not converge. If the margin of the input image is too small, we can pad some zeros around the image.

When the given image contains multiple objects, our model (\ref{eq:convhull2}) can also be used to compute the convex hull of each object separately. However, the value of $c$ should be chosen very carefully. If we want separated convex hulls, we should choose $c$ to be smaller than half of the distance between any two objects. Otherwise, we will get a big convex hull containing all the objects.  This idea will be explained again in Section \ref{sec5}.

\subsection{Convex hull model for images with outliers}
When the images contain noise and outliers, as the situation in Figure \ref{fig:noisy_hull_example}, it is inappropriate to require the model to enclose all the points into the convex hull. Instead, we can add a penalty term to the objective function. The new objective function is defined as
\begin{equation}
    F_2(x,\phi)=F_1(x,\phi)+\lambda(m(x)\phi(x))^+,\label{eq:F2}
\end{equation}
where $\lambda>0$ is a fixed parameter to be determined, $m(x)$ is the indicator function of $\Omega_1$, and $()^+$ is the positive part function:
\begin{equation}
    (y)^+=\begin{cases}
    y & y>0 \\
    0 & y\leq 0.
    \end{cases}
\end{equation}
Then the outliers problem can be written as
\begin{align}
    \underset{\phi}{\min}\quad &\int_{\Omega}F_2(x,\phi)dx\label{eq:convhull3}\tag{OC}\\
    {\rm subject\ to}\quad&|\nabla\phi(x)|=1 \text{ a.e. in }\Omega\\
    & \Delta\phi(x)\geq 0 \text{ a.e. in } \text{slev}_{\phi}^c.
\end{align}
The optimal solution of (\ref{eq:convhull3}) depends on the scale of the parameter $\lambda$. If $\lambda$ is too large, the boundary part of the object may be miss-excluded from the convex hull. If $\lambda$ is too small, some outliers or noise may also be included by mistake. We will elaborate more on the choice of $\lambda$ in the numerical experiments section.

\section{Numerical algorithm}\label{sec4}
In this section, we will introduce the ADMM-based algorithm for the two convex hull models. 
\subsection{Exact convex hull model}\label{sec4a}
First of all, we rewrite the model (\ref{eq:convhull2}) in discrete space, because the real image data is discrete. If we denote the discretized image domain by $\hat{\Omega}$, we can write the discretized exact model (\ref{eq:convhull2}) as
\begin{align}
    \underset{\phi}{\min}\quad &\sum_{x_i\in\hat{\Omega}}\tilde{F}_1(x_i,\phi)\label{eq:convhull2_discrete}\\
    {\rm subject\ to}\quad&|\nabla\phi(x_i)|=1,\ \forall\ x_i\in \hat{\Omega}\\
    & \Delta\phi(x_i)\geq 0,\ \forall\ x_i\in \text{slev}_{\phi}^c \label{eq:constraint_3a}\\
    & m(x_i)\phi(x_i)\leq 0,\ \forall\ x_i\in \hat{\Omega},
    \label{eq:constraint_4a}
\end{align}
where 
\begin{equation}
    \tilde{F}_1(x_i,\phi)=-\omega\phi(x_i)+\mu|\nabla H(\phi(x_i))|+\nu l(x_i)\phi(x_i)^2,
\end{equation}
and $l(x)$ here is the indicator function of the known boundary landmarks. The detailed discretization scheme is provided in Appendix \ref{appendix1}. Notice that the constraints (\ref{eq:constraint_3a}) and (\ref{eq:constraint_4a}) are required to hold at every pixel. Moreover, because the Heaviside function is discontinuous, in order to compute the derivative, we approximate it by 
\begin{align}
    H(s)\approx H_{\delta}(s)=\frac{1}{2}+\frac{1}{\pi}\arctan{\frac{s}{\delta}}.
\end{align}
where $\delta>0$ is a small number. 
To solve problem (\ref{eq:convhull2_discrete}), we introduce three auxiliary variables $z_1=\nabla\phi$, $z_2=\Delta\phi$ and $z_3=m\phi$. Then the problem is equivalent to
\begin{align}
    \underset{\phi}{\min}\quad &\sum_{x_i\in\hat{\Omega}}\tilde{F}_1(x_i,\phi)\label{eq:convhull4}\\
    {\rm subject\ to}\quad&|z_1(x_i)|=1,\ \forall\ x_i\in \hat{\Omega}\\
    & z_2(x_i)\geq 0,\ \forall\ x_i \in  \text{slev}_{\phi}^c \label{eq:constraint_c}\\
    & m(x_i)z_3(x_i)\leq 0,\ \forall\ x_i \in \hat{\Omega}\\
    & z_1=\nabla\phi,\ z_2=\Delta\phi,\ z_3=m\phi.
\end{align}
The associating augmented Lagrangian functional for (\ref{eq:convhull4}) is as follows:
\begin{align}
&L(\phi,z_1,z_2,z_3,\gamma_1,\gamma_2,\gamma_3)=\langle\gamma_1,\nabla\phi-z_1\rangle+\langle\gamma_2,\Delta\phi-z_2\rangle +\langle\gamma_3,m\phi-z_3\rangle\notag\\
&+\frac{\rho_1}{2}\Vert \nabla\phi-z_1\Vert_2^2+\frac{\rho_2}{2}\Vert \Delta\phi-z_2\Vert_2^2+\frac{\rho_3}{2}\Vert m\phi-z_3\Vert_2^2+\sum_{x_i\in\hat{\Omega}}\tilde{F}_1(\phi) \label{eq:auglag2}\\
&\text{s.t.} \ |z_1|=1 \text{ for } \forall\ x_i\in\hat{\Omega}, z_2\geq 0\text{ for } \forall\ x_i\in\text{slev}_{\phi}^c,\text{ and }z_3\leq 0\text{ for } \forall\ x_i \in \hat{\Omega}.
\end{align}
The inner product here is defined as 
\begin{equation}
\langle f,g\rangle=\sum_{x_i\in\hat{\Omega}}f(x_i)g(x_i) \text{ and } \Vert 
f\Vert_2^2=\langle f,f\rangle,
\end{equation}
when $f$ and $g$ are scalar-valued functions defined on $\Omega$, and
\begin{equation}
 \langle f,g\rangle=\sum_{x_i\in\hat{\Omega}}f(x_i)^Tg(x_i) \text{ and } \Vert f \Vert_2^2=\langle f,f\rangle,
\end{equation}
when $f$ and $g$ are vector-valued functions. solve (\ref{eq:auglag2}), we first choose an initial guess for $\phi^0$, and then update variables iteratively. If we denote the iteration number as $t$, the ADMM algorithm for (\ref{eq:auglag2})is derived as:
\begin{enumerate}
\item $z_1$ update:
\begin{align}
    z_1^{t+1}&=\underset{|z_1|=1}{\arg\min}\  L(\phi^t,z_1,z_2^t,z_3^t,\gamma_1^t,\gamma_2^t,\gamma_3^t) \notag\\
    &=\underset{|z_1|=1}{\arg\min}\  \langle\gamma_1^t,\nabla\phi^t-z_1\rangle+\frac{\rho_1}{2}\Vert\nabla\phi^t-z_1\Vert^2_2 \notag\\
    &=\underset{|z_1|=1}{\arg\min}\ \left\Vert z_1-\frac{\gamma_1^t}{\rho_1}-\nabla\phi^t \right\Vert^2_2\notag\\
    &=\frac{\frac{\gamma_1^t}{\rho_1}+\nabla\phi^t}{|\frac{\gamma_1^t}{\rho_1}+\nabla\phi^t|}.\label{z1update}
\end{align}

\item $z_2$ update:
\begin{align}
    z_2^{t+1}&=\underset{z_2\geq 0 \text{ in }\text{slev}_{\phi}^c}{\arg\min}\ L(\phi^t,z_1^{t+1},z_2,z_3^t,\gamma_1^t,\gamma_2^t,\gamma_3^t)\notag\\
    &=\underset{z_2\geq 0 \text{ in }\text{slev}_{\phi}^c}{\arg\min}\ \langle\gamma_2^t,\Delta\phi^t-z_2\rangle+\frac{\rho_2}{2}\Vert\Delta\phi^t-z_2\Vert^2_2\notag\\
    &=\underset{z_2\geq 0 \text{ in }\text{slev}_{\phi}^c}{\arg\min}\ 
    \left\Vert z_2-\frac{\gamma_2^t}{\rho_2}-\Delta\phi^t \right\Vert^2_2\notag\\
    &=\begin{cases}\text{max}\{0,\frac{\gamma_2^t}{\rho_2}+\Delta\phi^t\} & \phi(x)\leq c\\ \frac{\gamma_2^t}{\rho_2}+\Delta\phi^t & \phi(x)>c.\end{cases}
    \label{z2update}
\end{align}

\item $z_3$ update:
\begin{align}
    z_3^{t+1}&=\underset{z_3\leq 0}{\arg\min}\  L(\phi^t,z_1^{t+1},z_2^{t+1},z_3,\gamma_1^t,\gamma_2^t,\gamma_3^t) \notag\\
    &=\underset{z_3\leq 0}{\arg\min}\  \langle\gamma_3^t,m\phi^t-z_3\rangle+\frac{\rho_3}{2}\Vert m\phi^t-z_3\Vert^2_2 \notag\\
    &=\underset{z_3\leq 0}{\arg\min}\ \left\Vert z_3-\frac{\gamma_3^t}{\rho_3}-m\phi^t \right\Vert^2_2\notag\\
    &=\text{min}\left\{0,m\phi^t+\frac{\gamma_3^t}{\rho_3}\right\}.\label{z3update}
\end{align}

\item $\phi$ update:
\begin{align}
    \phi^{t+1}&=\underset{\phi}{\arg\min}\  L(\phi,z_1^{t+1},z_2^{t+1},z_3^{t+1},\gamma_1^t,\gamma_2^t,\gamma_3^t) \notag\\
    &=\underset{\phi}{\arg\min}\  \sum_{x_i\in\hat{\Omega}} \tilde{F}_1(\phi)\notag\\
    &+\langle\gamma_1^t,\nabla\phi-z_1^{t+1}\rangle+\langle\gamma_2^t,\Delta\phi-z_2^{t+1}\rangle\notag\\ &+\langle\gamma_3^t,m\phi-z_3^{t+1}\rangle+\frac{\rho_1}{2}\Vert\Delta\phi-z_1^{t+1}\Vert^2_2\notag\\
    &+\frac{\rho_2}{2}\Vert\nabla\phi-z_2^{t+1}\Vert^2_2+\frac{\rho_3}{2}\Vert m\phi-z_3^{t+1}\Vert^2_2. \label{phifunctional}
\end{align}
Using the calculus of variations, we can see that the optimal $\phi$ must satisfy the periodic boundary condition and the following equality
\begin{align}
    &&-{\divg}(\gamma_1^t+\rho_1(\nabla\phi-z_1^{t+1}))+\Delta(\gamma_2^t+\rho_2(\Delta\phi-z_2^{t+1}))\notag\\
    &&+m(\gamma_3+\rho_3(\phi-z_3^{t+1}))+\tilde{F}'_1(\phi)=0.
\end{align}
In order to solve this equation efficiently, we only keep $\Delta\phi$ and $\Delta^2\phi$ on the left-hand side and move the others to the right hand. Approximately, the $\phi$ update can be written as:
\begin{align}
     &-\rho_1\Delta\phi^{t+1}+\rho_2\Delta^2\phi^{t+1}\\
    &={\divg}(\gamma_1^t-\rho_1 z_1^{t+1})-\Delta(\gamma_2^t-\rho_2 z_2^{t+1})\notag\\
    &-(\gamma_3+\rho_3(\phi^t-z_3^{t+1}))m-\tilde{F}'_1(\phi^t).\notag\label{eq:phi_4th_pde}
\end{align}
Here we use the same technique with \cite{luo2018convex} to solve this fourth order equation efficiently. By adding a proximity term $\frac{\rho_0}{2}\Vert\phi-\phi^t\Vert_2^2$ to (\ref{eq:auglag2}), the $\phi$ update can be rewritten as
\begin{align}
    & -\rho_1\Delta\phi^{t+1}+\rho_2\Delta^2\phi^{t+1}+\rho_0\phi^{t+1}\label{eq:phi_rhs}
    \\
    &=-\Delta(\gamma_2^t-\rho_2 z_2^{t+1})+{\divg}(\gamma_1^t-\rho_1 z_1^{t+1})\notag\\
    &-(\gamma_3+\rho_3(\phi^t-z_3^{t+1}))m-\tilde{F}'_1(\phi^t)+\rho_0\phi^t.\notag
\end{align}
If we denote the right-hand side of (\ref{eq:phi_rhs}) as $g_1(\phi^t)$, and choose $\rho_1=2\sqrt{\rho_0\rho_2}$, we can split the original PDE (\ref{eq:phi_rhs}) into two second order PDEs:
\begin{align}
    \begin{cases}
    (\sqrt{\rho_2}\Delta-\sqrt{\rho_0})\psi^{t+1}=g_1(\phi^t)\\
     (\sqrt{\rho_2}\Delta-\sqrt{\rho_0})\phi^{t+1}=\psi^{t+1}.
    \end{cases}\label{eq:phi_2nd_pde}
\end{align}
The equations for $\phi^{t+1}$ and $\psi^{t+1}$ have exactly the same structure and satisfy the same boundary condition so they can be solved by the same algorithm. In \cite{luo2018convex}, the authors use the discrete cosine transform (DCT) to solve the equations. Since in this paper, we assume the image is periodic, $\phi^{t+1}$ and $\psi^{t+1}$ satisfy the periodic boundary condition, we apply the fast Fourier transform (FFT) to solve the problem instead of DCT. If we denote the FFT and its inverse as $\mathcal{F}()$ and $\mathcal{F}^{-1}()$, then after applying FFT on both sides of (\ref{eq:phi_2nd_pde}), we have
\begin{align}
    \begin{cases}
    \mathcal{F}(\psi^{t+1})(i,j)=\mathcal{F}(g_1(\phi^t))(i,j)/c(i,j)\\
    \mathcal{F}(\phi^{t+1})(i,j)=\mathcal{F}(\psi^{t+1})(i,j)/c(i,j),
    \end{cases}\label{eq:phi_fft}
\end{align}
where $c(i,j)$ equals to
\begin{equation}
    2\sqrt{\rho_2}(\cos(\frac{2\pi(i-1)}{M})+\cos(\frac{2\pi(j-1)}{N})-2)-\sqrt{\rho_0}.
\end{equation}
Then $\phi^{t+1}$ can be obtained by $\mathcal{F}^{-1}(\mathcal{F}(\phi^{t+1}))$.

\item $\gamma$ update:
For $\gamma_1,\gamma_2$ and $\gamma_3$, we do the following updates:
\begin{align}
    &\gamma_1^{t+1}=\gamma_1^{t}+\rho_1(\nabla\phi^{t+1}-z_1^{t+1}),\label{gamma1}\\
    &\gamma_2^{t+1}=\gamma_2^{t}+\rho_2(\Delta\phi^{t+1}-z_2^{t+1})\label{gamma2},\\
    &\gamma_3^{t+1}=\gamma_3^{t}+\rho_3(m\phi^{t+1}-z_3^{t+1})\label{gamma3}.
\end{align}
\end{enumerate}
We summarize this convex hull algorithm as Algorithm \ref{exactalgo}.
\begin{algorithm}
\caption{Exact convex hull algorithm}
\begin{algorithmic}[1]
\renewcommand{\algorithmicrequire}{\textbf{Input:}}
\renewcommand{\algorithmicensure}{\textbf{Output:}}
\REQUIRE A binary image $I$, $\rho_0$ and $\rho_2$, maximum number of iteration $M$, and a threshold $\epsilon>0$.
\ENSURE  $\phi$
\\ \textit{Initialization} :
Let $\phi^1$ to be the SDF of an initial shape and $\phi^0=0$. Set the largest iteration number $M$ and $t=1$.
\WHILE{$t<M$ \& $\frac{1}{|\Omega|}\int_{\Omega}|\phi^{t}(x)-\phi^{t-1}(x)|dx>\epsilon$}
\STATE update $z_1$ by (\ref{z1update})
\STATE update $z_2$ by (\ref{z2update})
\STATE update $z_3$ by (\ref{z3update})
\STATE update $\phi$ by solving (\ref{eq:phi_2nd_pde})
\STATE update $\gamma_1,\gamma_2$ and $\gamma_3$ by (\ref{gamma1}), (\ref{gamma2}) and (\ref{gamma3})
\STATE $t=t+1$
\ENDWHILE
\RETURN $\{x|\phi(x)\leq 0\}$ 
\end{algorithmic} 
\label{exactalgo}
\end{algorithm}

\subsection{Convex hull model for images with outliers}\label{sec4b}
For the outliers model (\ref{eq:convhull3}), we also write it in discrete form:
\begin{align}
    \underset{\phi }{\min}\quad &\sum_{x_i\in\hat{\Omega}}\tilde{F}_2(x_i,\phi)\label{eq:convhull3_discrete}\\
    {\rm subject\ to}\quad&|\nabla\phi(x_i)|=1, \text{ for }\forall\ x_i \in  \hat{\Omega}\\
    & \Delta\phi(x_i)\geq 0,\text{ for }\forall\ x_i \in \text{slev}_{\phi}^c,
\end{align}
where $\tilde{F}_2(\phi)=\tilde{F}_1(\phi)+\lambda (m\phi)^+$. Similarly, we also introduce two auxiliary variable $z_1$ and $z_2$ such that $z_1=\nabla\phi$ and $z_2=\Delta\phi$. Then the augmented Lagrangian functional is
\begin{align}
&L(\phi,z_1,z_2,z_3,\gamma_1,\gamma_2,\gamma_3)=\langle\gamma_1,\nabla\phi-z_1\rangle+\langle\gamma_2,\Delta\phi-z_2\rangle\notag \\
&+\frac{\rho_1}{2}\Vert \nabla\phi-z_1\Vert_2^2+\frac{\rho_2}{2}\Vert \Delta\phi-z_2\Vert_2^2
+\sum_{x_i\in\hat{\Omega}}\tilde{F}_2(\phi) \label{eq:auglag1}\\
&\text{s.t.} \ |z_1|=1\text{ in }\hat{\Omega},\ z_2\geq 0 \text{ for }\forall\  x_i\text{ in }\text{slev}_{\phi}^c.
\end{align}
The ADMM updates for $z_1$ and $z_2$ are the same with (\ref{z1update}) and (\ref{z2update}). The $\phi$ update is also very similar. Using the same idea, we can split it into two second-order PDEs:
\begin{align}
    \begin{cases}
    (\sqrt{\rho_2}\Delta-\sqrt{\rho_0})\psi^{t+1}=g_2(\phi^t),\\
     (\sqrt{\rho_2}\Delta-\sqrt{\rho_0})\phi^{t+1}=\psi^{t+1},
    \end{cases}\label{eq:phi_2nd_pde2}
\end{align}
where $g_2(\phi^t)$ equals to
\begin{align}
    -\Delta(\gamma^t_2-\rho_2 z_2^{t+1})+{\divg}(\gamma_1^t-\rho_1z_1^{t+1})-\tilde{F}_2(\phi)+\rho_0 \phi^t.
\end{align}
This system can also be solved by FFT in the same way with (\ref{eq:phi_2nd_pde}).
We summarize the outliers algorithm in Algorithm \ref{algo:outliers}.
\begin{algorithm}
\caption{Convex hull algorithm for image with outliers}
\begin{algorithmic}[1]
\renewcommand{\algorithmicrequire}{\textbf{Input:}}
\renewcommand{\algorithmicensure}{\textbf{Output:}}
\REQUIRE A binary image $I$, $\rho_0$ and $\rho_2$, maximum number of iteration $M$, and a threshold $\epsilon>0$.
\ENSURE  $\phi$
\\ \textit{Initialization} :
Let $\phi^1$ to be the SDF of an initial shape and $\phi^0=0$. Set $t=1$.
\WHILE{$t<M$ \& $\frac{1}{|\Omega|}\int_{\Omega}|\phi^{t}(x)-\phi^{t-1}(x)|dx>\epsilon$}
\STATE update $z_1$ by (\ref{z1update})
\STATE update $z_2$ by (\ref{z2update})
\STATE update $\phi$ by solving (\ref{eq:phi_2nd_pde2})
\STATE update $\gamma_1$ and $\gamma_2$ by (\ref{gamma1}) and (\ref{gamma2})
\STATE $t=t+1$
\ENDWHILE
\RETURN $\{x|\phi(x)\leq 0\}$ 
\end{algorithmic} 
\label{algo:outliers}
\end{algorithm}

\section{Numerical experiments}\label{sec5}
In this section, we will conduct some tests for both Algorithm \ref{exactalgo} and Algorithm \ref{algo:outliers}. For the exact model, we test it on many objects with different shapes and compare our results with the quickhull algorithm. For the outliers model, we test it on some binary images with either randomly generated outliers or real outliers. The results show that our exact model can find the convex hull with very small error and our outliers model can filter out the outliers accurately. In the case that multiple objects are given, our methods also work well.

\subsection{Exact convex hull model}
\begin{figure}
    \centering
    \subfloat[Eggs]{\includegraphics[width=3cm]{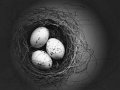}}\ 
    \subfloat[Frog]{\includegraphics[width=3cm]{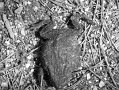}}\ 
    \subfloat[Helicopter]{\includegraphics[width=3cm]{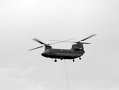}}\\
    \subfloat[Moth]{\includegraphics[width=3cm]{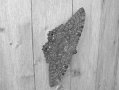}}\ 
    \subfloat[Tendrils]{\includegraphics[width=3cm]{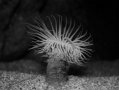}}\ 
    \subfloat[Owl]{\includegraphics[width=3cm]{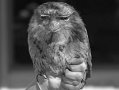}}\\
    \subfloat[Boat]{\includegraphics[width=3cm]{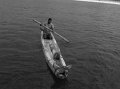}}\ 
    \subfloat[Castle]{\includegraphics[width=3cm]{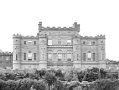}}\ 
    \subfloat[Cart]{\includegraphics[width=3cm]{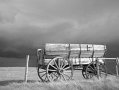}}\\
    \caption{Original images}
    \label{original_images}
\end{figure}

\begin{figure}
    \centering
    \subfloat[Eggs]{\includegraphics[width=3cm]{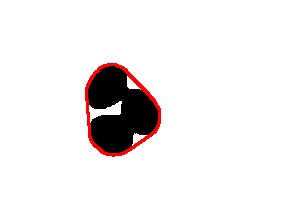}}
    \subfloat[Frog]{\includegraphics[width=3cm]{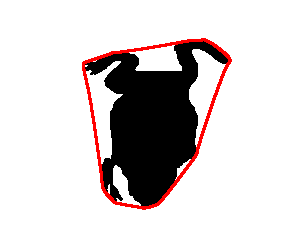}}
    \subfloat[Helicopter]{\includegraphics[width=3cm]{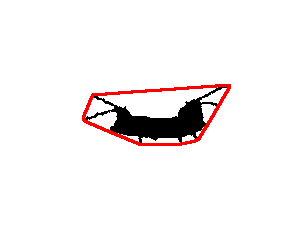}}\\
    \subfloat[Moth]{\includegraphics[width=3cm]{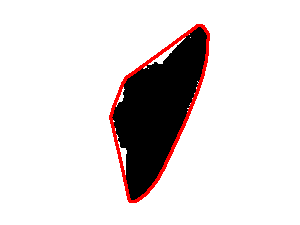}}
    \subfloat[Tendrils]{\includegraphics[width=3cm]{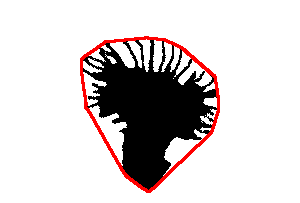}}
    \subfloat[Owl]{\includegraphics[width=3cm]{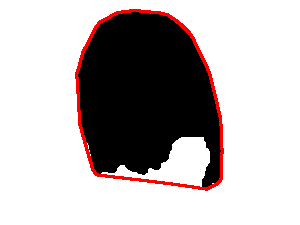}}\\
    \subfloat[Boat]{\includegraphics[width=3cm]{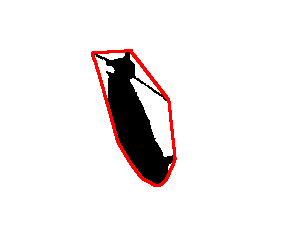}}
    \subfloat[Castle]{\includegraphics[width=3cm]{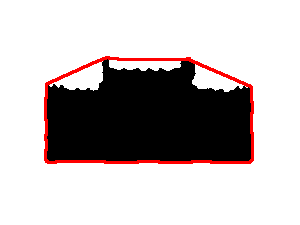}}
    \subfloat[Cart]{\includegraphics[width=3cm]{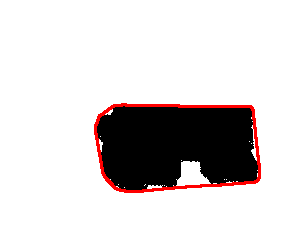}}\\
    \caption{Convex hulls found by Algorithm \ref{exactalgo}.}
    \label{exacthullimg}
\end{figure}

\begin{figure}
    \centering
    \subfloat[]{\includegraphics[width=3cm]{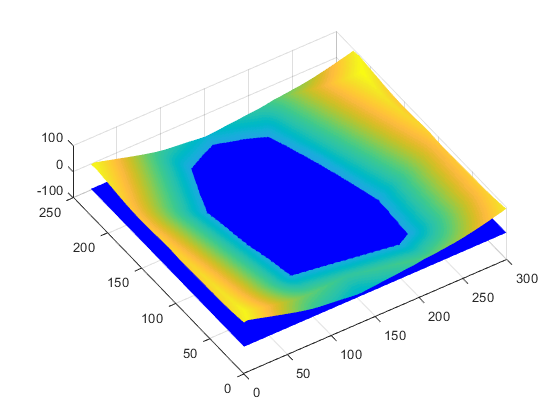}}\quad
    \subfloat[]{\includegraphics[width=3cm]{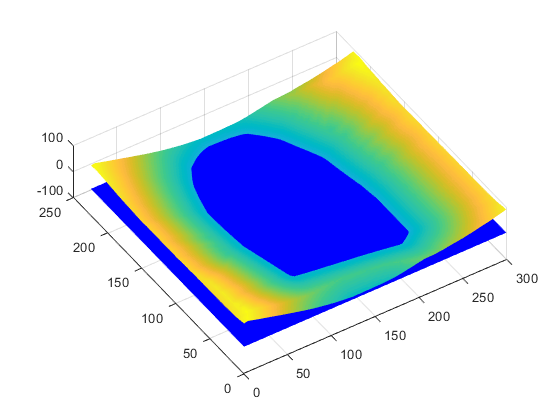}}\quad
    \subfloat[]{\includegraphics[width=3cm]{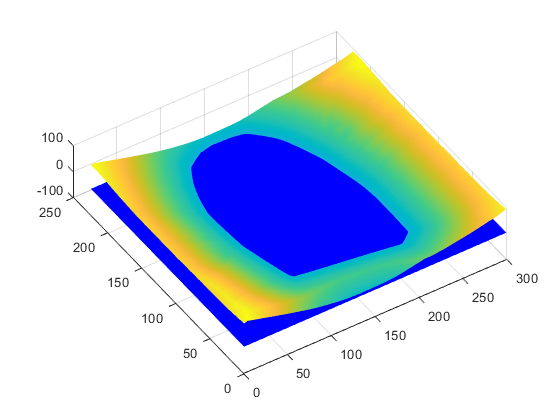}}\\
    \subfloat[]{\includegraphics[width=3cm]{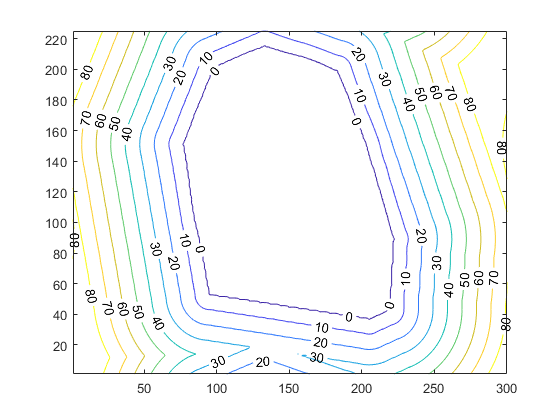}}\quad
    \subfloat[]{\includegraphics[width=3cm]{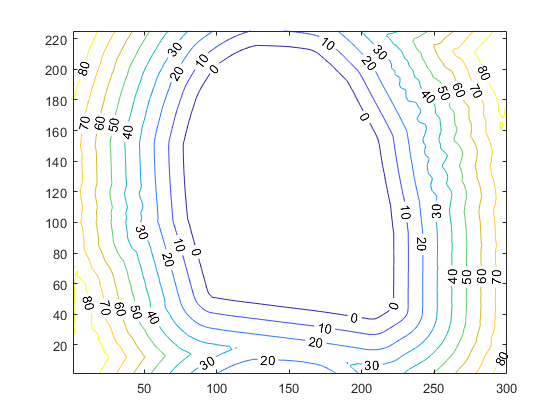}}\quad
    \subfloat[]{\includegraphics[width=3cm]{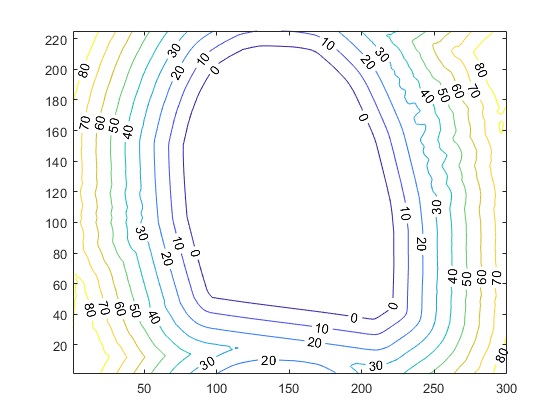}}\\
    \caption{(a), (b) and (c) plot the values of function $\phi$ at the 0th, 200th and 400th iterations. (d), (e) and (f) plot the corresponding level-set curves of $phi$ at the 0th, 200th and 400th iterations.}
    \label{fig:evolution_level_set}
\end{figure}

In this part, we choose 9 pictures from \cite{AlpertGBB07}. The original images are listed in Figure \ref{original_images}, and the convex hulls yielded by Algorithm \ref{exactalgo} using the provided ground truth mask are shown in Figure \ref{exacthullimg}. To measure the accuracy of our proposed algorithm, we compare our results against the quickhull algorithm using the relative distance error \cite{rufai2015convex}: 
\begin{align}
    &\text{err}(C_2)=\frac{\text{dist}_H(C_1,C_2)}{D(C_1)} \label{errmeasure}\\
    &D(C_1)=2\sqrt{\frac{\text{area}(C_1)}{\pi}},
\end{align}
where $C_2$ is the convex hull found by our proposed algorithm, $C_1$ is the convex hull found by the benchmark algorithm, i.e, quickhull, and $\text{dist}_H$ is the Hausdorff distance:
\begin{align}
    \text{dist}_H(C_1,C_2)=&\max\{\sup\{\text{dist}(x,C_2)|x\in C_1\},\notag\\
    &\quad\quad \sup\{\text{dist}(y,C_1)|y\in C_2\}\}.
\end{align}
The relative distance errors are shown in Table \ref{exacterror}. From the results, we can see that our proposed algorithm can yield the convex hull of given objects with very small error, and the convexity of the region is also guaranteed. The error of all the images are under $2\%$. In this set of experiment, we use the same set of parameters for all images: $\rho_0=1$, $\rho_2=15$, $\rho_3=1$, $\omega=0.01$, $\mu=5$, $\nu=10$ and $c=20$. Actually, this set of parameters is very robust to various images, which means we don't need to tune the parameters when applying it to most of images. In Figure \ref{fig:evolution_level_set}, we show the evolution of the SDF of the owl image. After only hundreds of iterations, the SDF of the convex hull can be found accurately, and the level-set curves up to 20 are all convex.

\begin{table}
\caption{The relative errors of Algorithm \ref{exactalgo}}
\centering
\begin{tabular}{@{}cccccc@{}}
\toprule
name  & \textbf{Eggs}    & \textbf{Frog} & \textbf{Helicopter}   & \textbf{Moth} & \textbf{Tendrils} \\ \midrule
error & 1.28\%                & 1.51\%        & 1.13\%          & 0.92\%        & 0.73\%          \\\midrule
name  & \textbf{Owl} & \textbf{Boat}  & \textbf{Castle} & \textbf{Cart} &                 \\\midrule
error & 0.61\%               & 1.08\%        & 0.63\%          & 0.79\%        &                 \\ \bottomrule
\end{tabular}
\label{exacterror}
\end{table}

\begin{figure}
    \centering
    \subfloat[]{\includegraphics[width=3cm]{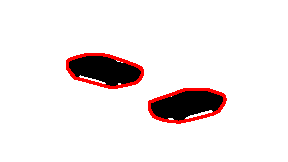}}
    \subfloat[]{\includegraphics[width=3cm]{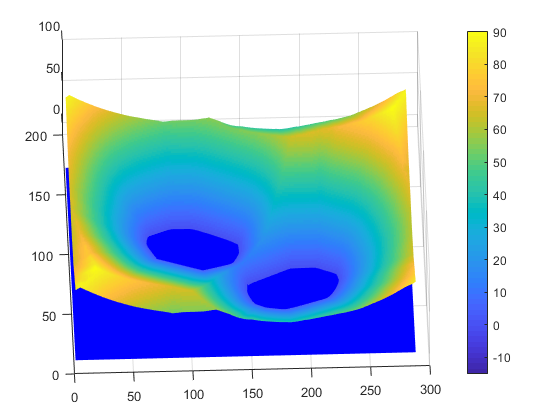}}
    \subfloat[]{\includegraphics[width=3cm]{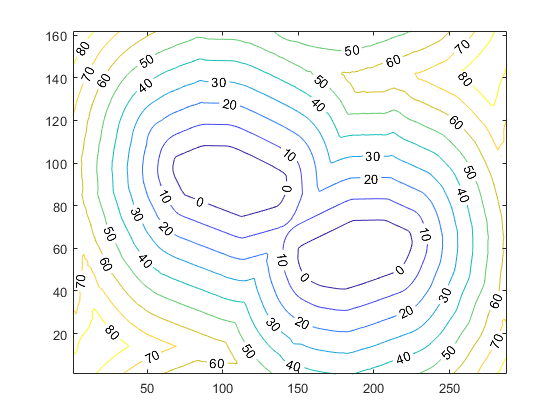}}
    \subfloat[]{\includegraphics[width=3cm]{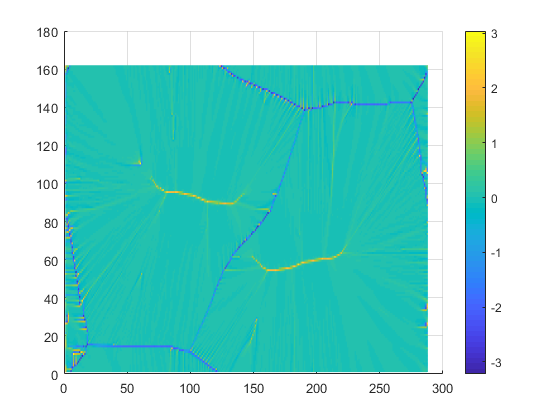}}\\
    \subfloat[]{\includegraphics[width=3cm]{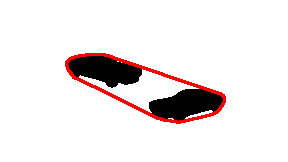}}
    \subfloat[]{\includegraphics[width=3cm]{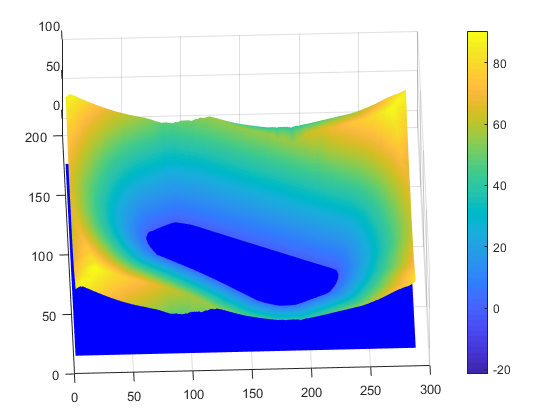}}
    \subfloat[]{\includegraphics[width=3cm]{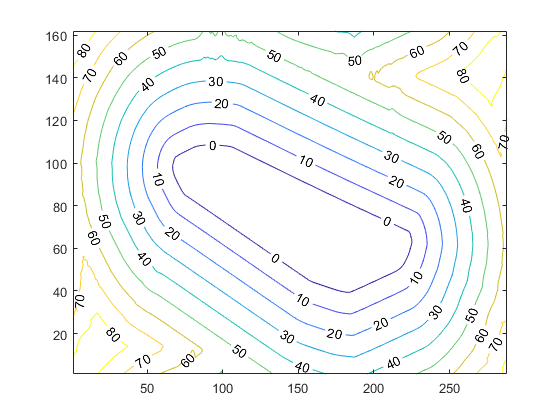}}
    \subfloat[]{\includegraphics[width=3cm]{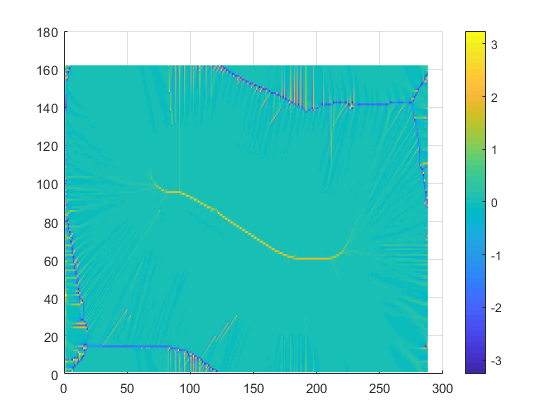}}
    \caption{(a) shows the convex hulls result when $c=10$. (b), (c) and (d) show the corresponding function values of $\phi$, level-set curves of $\phi$ and value of $\Delta\phi$. (e) shows the convex hulls result when $c=30$. (f), (g) and (h) show the corresponding function values of $\phi$, level-set curves of $\phi$ and value of $\Delta\phi$.}
    \label{fig:level_set_plot}
\end{figure}

\begin{figure}
    \centering
    \subfloat[Pokemon]{\includegraphics[width=3cm]{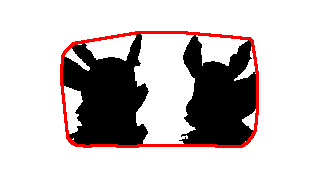}}
    \subfloat[Pokemon]{\includegraphics[width=3cm]{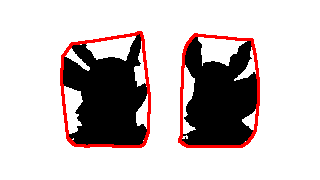}}
    \subfloat[Birds]{\includegraphics[width=3cm]{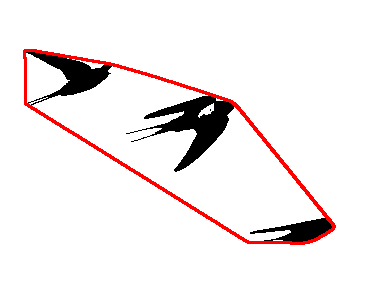}}
    \subfloat[Birds]{\includegraphics[width=3cm]{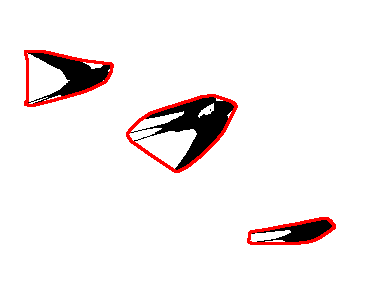}}\\
    \caption{Convex hulls of multi-objects by chosing proper values of $c$. }
    \label{fig:multihull}
\end{figure}

\begin{figure}
    \centering
    \subfloat[Original image]{\includegraphics[width=3cm]{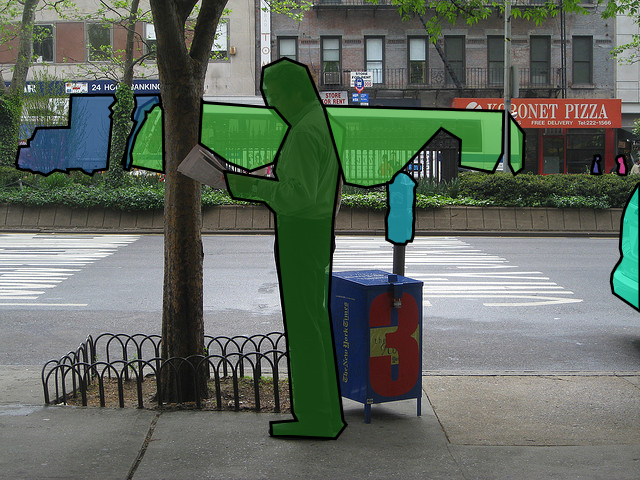}}
    \subfloat[Convex hull of a bus]{\includegraphics[width=3cm]{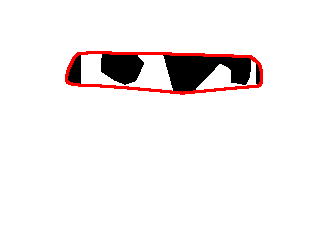}}
    \subfloat[Original image]{\includegraphics[width=3cm]{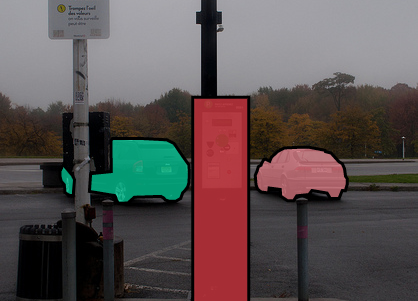}}
    \subfloat[Convex hulls of two cars]{\includegraphics[width=3cm]{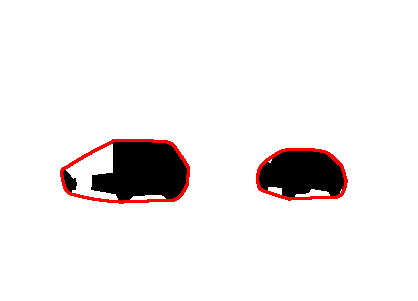}}\\
    \caption{Convex hulls in object detecting tasks.}
    \label{fig:car_example}
\end{figure}

As we mentioned before, the algorithm can be used to compute the convex hulls of multiple objects, and the key is to properly choose the value of $c$ in (\ref{eq:constraint_4}). To better explain this, we use an image of two cars as an example. In Figure \ref{fig:level_set_plot}, the subfigure (a) shows the convex hulls of two cars separately, and (b) plots its corresponding SDF. If we loot at the level-set curves of the SDF (c), we can see that only the 0 and 10 level-set curves are convex. From the plot of Laplacian $\phi$ (d), we can also observe that $\Delta\phi$ is negative on a skeleton between the two objects. In model (\ref{eq:convhull2}), if we require $\Delta\phi\geq 0$ in $\text{slev}_{\phi}^c$ for some $0\leq c\leq 10$, this SDF  minimizes the energy functional and produces convex hulls for two separated objects. However, if we want a big convex hull containing both objects, just like (e), we can choose a large $c$, for example, $c=30$. In this case, the SDF in (b) no longer satisfies the constraint $\Delta\phi\geq 0$ in the subregion $\text{slev}_{\phi}^c$, since its 20 and 30 level-set curves are not convex. Instead, our algorithm will return the SDF of the big convex hull (f). We can observe that its level-set curves (g) are convex for $c$ up to 30, and $\Delta\phi$ (h) is always non-negative in the middle. In fact, as long as the $c$ is smaller than half of the distance between the two objects, we will obtain two separated convex hulls. More numerical examples are shown in Figure \ref{fig:multihull}. We also apply the algorithm to some images from the COCO dataset \cite{lin2014microsoft} in Figure \ref{fig:car_example}. Even though the objects in the images are partially blocked, the whole object can still be identified via our algorithm.

\subsection{Convex hull model for images with outliers}
\begin{figure}
    \centering
    \subfloat[Eggs]{\includegraphics[width=3cm]{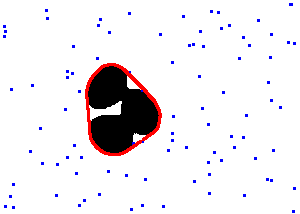}}\ \ 
    \subfloat[Frog]{\includegraphics[width=3cm]{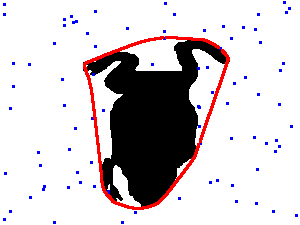}}\ \  
    \subfloat[Helicopter]{\includegraphics[width=3cm]{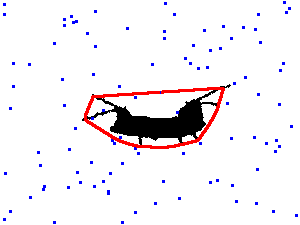}}\\
    \subfloat[Moth]{\includegraphics[width=3cm]{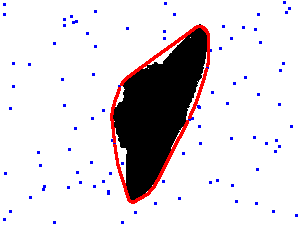}}\ \  
    \subfloat[Tendrils]{\includegraphics[width=3cm]{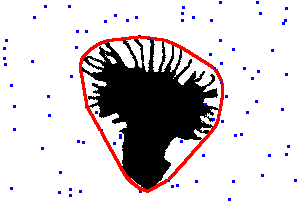}}\ \  
    \subfloat[Owl]{\includegraphics[width=3cm]{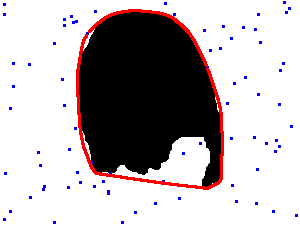}}\\
    \subfloat[Boat]{\includegraphics[width=3cm]{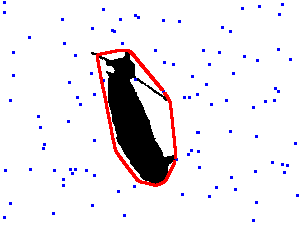}}\ \  
    \subfloat[Castle]{\includegraphics[width=3cm]{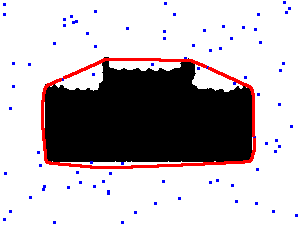}}\ \  
    \subfloat[Cart]{\includegraphics[width=3cm]{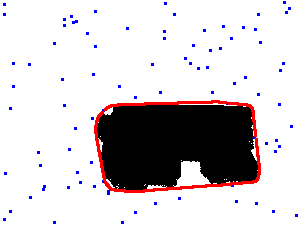}}\\
    \caption{Images with outliers and their convex hulls.}
    \label{approxhullimg}
\end{figure}

\begin{figure}
    \centering
    \subfloat[]{\includegraphics[width=3cm]{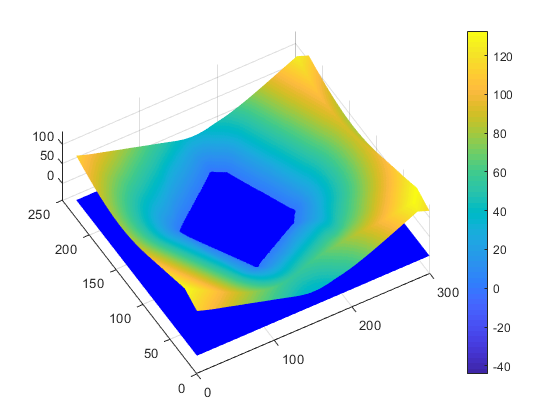}}
    \subfloat[]{\includegraphics[width=3cm]{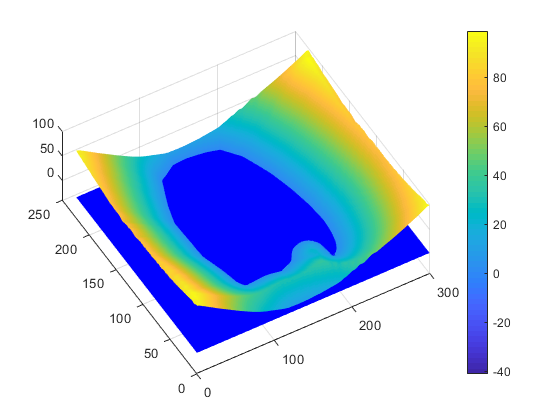}}
    \subfloat[]{\includegraphics[width=3cm]{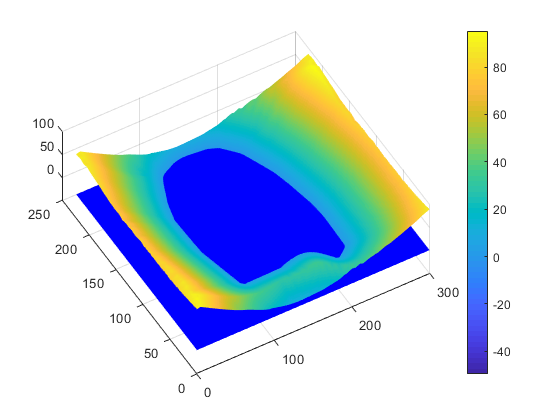}}
    \subfloat[]{\includegraphics[width=3cm]{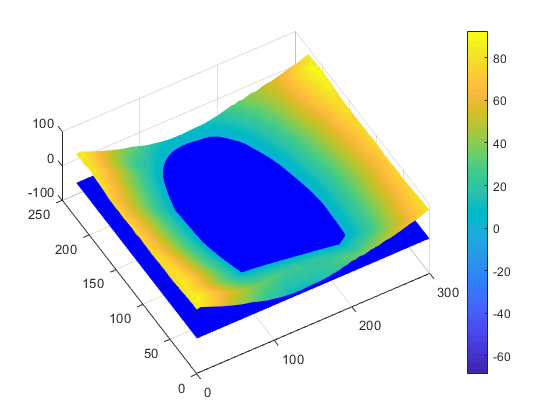}}
    \\
    \subfloat[]{\includegraphics[width=3cm]{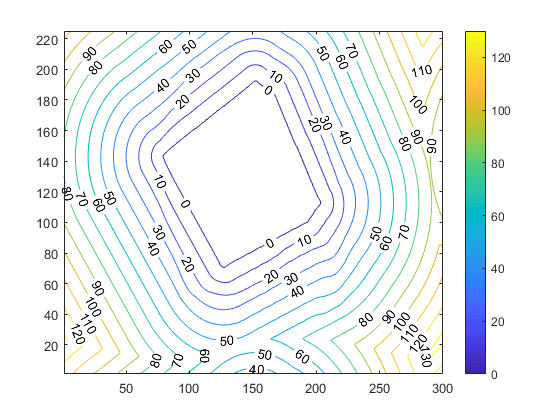}}
    \subfloat[]{\includegraphics[width=3cm]{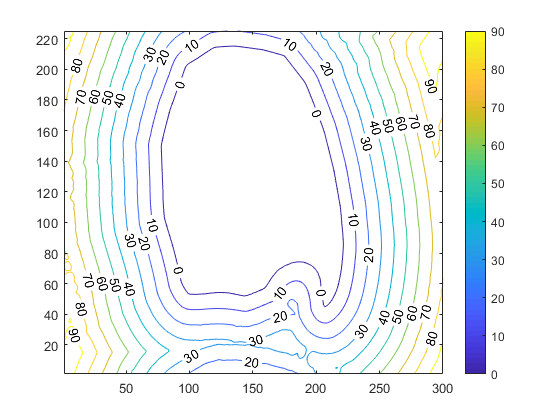}}
    \subfloat[]{\includegraphics[width=3cm]{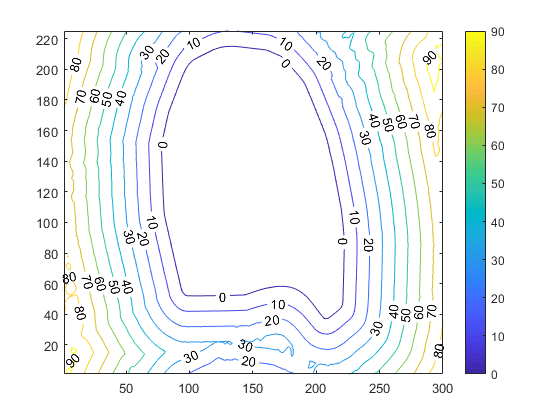}}
    \subfloat[]{\includegraphics[width=3cm]{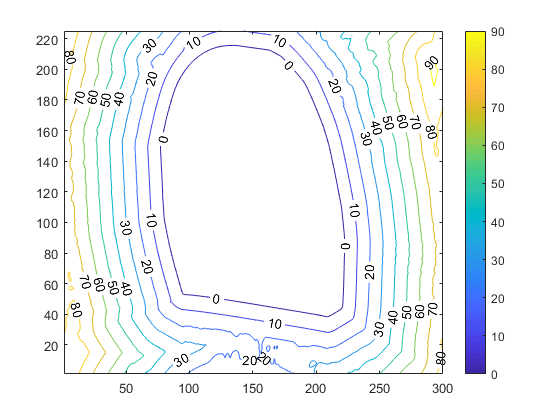}}\\
    \caption{(a), (b), (c) and (d) plot the function value of $\phi$ at the 0th, 200th, 800th and 3200th iterations when computing the convex hull of the owl image. (e), (f), (g) and (h) show the corresponding level-set curves of $\phi$ at the 0th, 200th, 800th and 3200th iterations.}
    \label{fig:evolution_level_set_inexact}
\end{figure}

For the outliers model (\ref{eq:convhull3}), we don't require the curve to enclose all the given points. Instead, we allow our model to exclude some outliers when it is too "expensive" to enclose them. In this case, the choice of parameter is very important. When $\lambda$ in (\ref{eq:F2}) is too large, the algorithm may leave some part of the object outside the convex hull, if $\lambda$ is too small, the convex hull may enclose some outliers by mistake. 
To test our algorithm, we first add some random noise to the images in Figure \ref{exacthullimg}. The approximated convex hulls by Algorithm \ref{algo:outliers} are shown in Figure \ref{approxhullimg}. We also compute the relative error using (\ref{errmeasure}), and the errors are listed in Table \ref{inexacterror}. From the results, we see that our proposed model can correctly filter out most of outliers and find the convex hull with small error. Though the error is larger than Table \ref{exacterror}, it is still acceptable. The largest error occurs for the helicopter image. It has a very long and thin blade on the top.  The cost is very high to include the whole blade. For the helicopter and boat image, we use: $\rho_0=1$, $\rho_2=20$, $\omega=0.005$, $\mu=3$, $\nu=20$, $\lambda=4$. For the rest of the images, we set $\lambda=3$ and other parameters remain the same. We want to emphasis here that the proposed algorithm is rather stable with these parameters. Essentially, we can use the same values of the parameters for different images with similar image size and noise level. To increase the stability of the algorithm, we also pad some zeros around the input image. We also plot the evolution of the SDF of the owl image in Figure \ref{fig:evolution_level_set_inexact}, where we can see the algorithm find the optimal solution in about 3000 iterations.

\begin{table}
\caption{The Relative Errors of Algorithm \ref{algo:outliers}.}
\centering
\begin{tabular}{@{}cccccc@{}}
\toprule
name  & \textbf{Eggs}    & \textbf{Frog} & \textbf{Helicopter}   & \textbf{Moth} & \textbf{Tendrils} \\ \midrule
error & 1.28\%                & 4.79\%        & 9.63\%          & 3.33\%        & 4.39\%          \\\midrule
name  & \textbf{Owl} & \textbf{Boat}  & \textbf{Castle} & \textbf{Cart} &                 \\\midrule
error & 2.73\%               & 6.85\%        & 3.79\%          & 3.93\%        &                 \\ \bottomrule
\end{tabular}
\label{inexacterror}
\end{table}

\begin{figure}
    \centering
    \includegraphics[width=3cm]{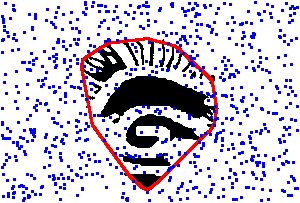}\quad
    \includegraphics[width=3cm]{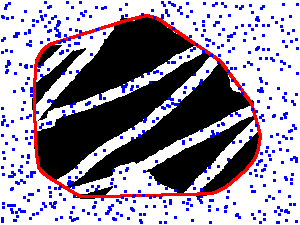}
    \caption{Convex hulls of occluded objects with outliers.}
    \label{fig:occluded}
\end{figure}

\begin{figure}
    \centering
    \subfloat[$\lambda=1$]{\includegraphics[width=3cm]{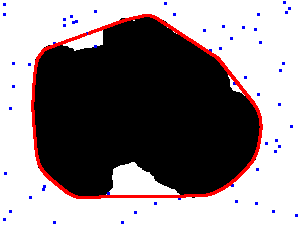}}\quad 
    \subfloat[$\lambda=2$]{\includegraphics[width=3cm]{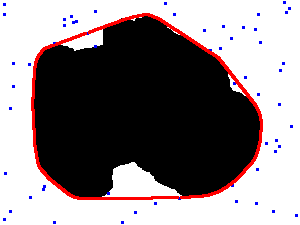}}\quad
    \subfloat[$\lambda=3$]{\includegraphics[width=3cm]{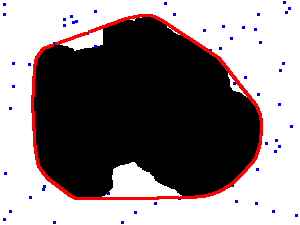}}
    \caption{Convex hulls of camera with different $\lambda$.}
    \label{fig:camera_lambda}
\end{figure}

\begin{figure}
    \centering
    \subfloat[$\lambda=3$]{\includegraphics[width=3cm]{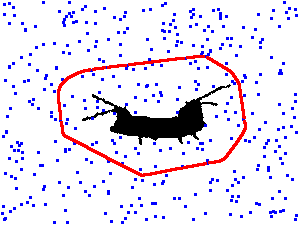}}\quad 
    \subfloat[$\lambda=2$]{\includegraphics[width=3cm]{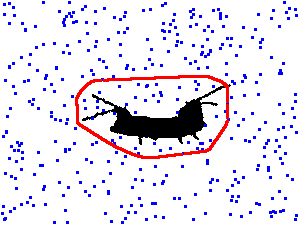}}\\
    \subfloat[$\lambda=1$]{\includegraphics[width=3cm]{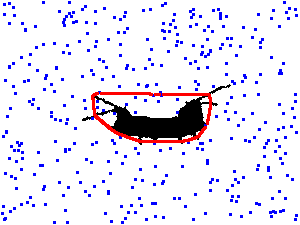}}\quad
    \subfloat[$\lambda=1$ with landmarks]{\includegraphics[width=3cm]{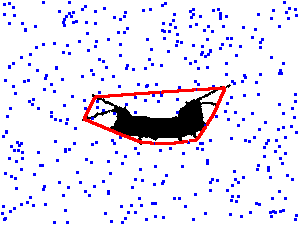}}\\
    \caption{Convex hulls of helicopter with different $\lambda$ and landmarks.}
    \label{fig:tune_lambda}
\end{figure}

Experiments on more challenging examples are also conducted. In Figure \ref{fig:occluded}, we manually erase part of the objects and add some salt and pepper noise. Our algorithm is still able to find the convex hulls accurately. To see if our algorithm is robust to the change of parameters, we conduct another two sets of examples. The first one is a camera image with a low level of noise (Figure \ref{fig:camera_lambda}) and the second one is a helicopter image with a high level of noise (Figure \ref{fig:tune_lambda}). From the results, we can make the following observations. First, when the shape of the object is simple and the amount of outliers is moderate, our algorithm is very robust to the change of $\lambda$. Second, under the existence of a large number of outliers, we should choose small $\lambda$ to get the accurate convex hull. Third, for the objects containing protruding parts, like the blades of the helicopter, the algorithm may leave parts of the object outside the convex hull, because it is too expensive to enclose the whole object. Finally, to further improve the accuracy, we can add some boundary landmarks to help locate the boundary of convex hulls. In the last subfigure of Figure \ref{fig:tune_lambda}, we add two landmarks on the end of blades (marked by red dots) and we find the result is significantly improved.

\subsection{Convex hulls for sets of isolated points}
\begin{figure}[ht]
    \centering
    \subfloat[Exact convex hull ]{\includegraphics[width=3cm]{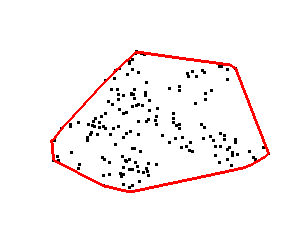}}\quad
    \subfloat[Convex hull with outliers]{\includegraphics[width=3cm]{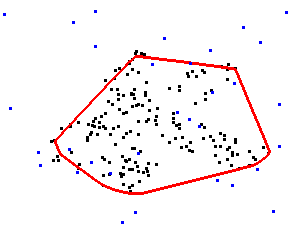}}\quad
    \caption{Convex hulls of a set of isolated points.}
    \label{fig:isolated}
\end{figure}

Our model works not only for a connected region, but also for sets of isolated points. To illustrate this, we generate random data points in a star fish mask, and then apply Algorithm \ref{exactalgo} and \ref{algo:outliers} on it. The results are shown in Figure \ref{fig:isolated}, where the given points are marked in black and noises are marked in blue. One can see that even if the given point set is not a connected region, our algorithms are also able to find the convex hulls accurately. 

\section{Conclusion}\label{sec6}
The convex hull problems arise from various areas. In some applications of image processing, convex hulls of objects need to be computed. In this work, we present some variational models for convex hull problem using level-set representation method. 
This method not only can handle the traditional convex hull problems, but also can tackle 
the convex hull problem for multiple objects and object(s) with outliers, where all the conventional methods fail
The numerical experiments show the proposed methods can identify the convex hulls with high accuracy. 
In the future, we will study on more efficient algorithm for the proposed models.

\appendix
\section{Discretization scheme of the algorithm}\label{appendix1}
Suppose our given image is of size $M\times N$, then we denote the discretized image domain as $\hat{\Omega}=\{(m,n)|m=1,\dots,M,n=1,\dots,N\}$. For any feasible solution to (\ref{eq:auglag1}) or (\ref{eq:auglag2}) defined on $\hat{\Omega}$, we require $\phi$ satisfy the periodic boundary condition, so we should extend $\phi$ as follow:
\begin{align}
&\phi(0,n)=\phi(M,n),\phi(M+1,n)=\phi(1,n),\\
&\phi(m,0)=\phi(m,N),\phi(m,N+1)=\phi(m,1),
\end{align}
where $m=1,\dots,M$ and $n=1,\dots,N$. With this extension, we can define our gradient operator using forward difference:
\begin{align}
  \nabla\phi(m,n)=\begin{bmatrix} \partial^+_x\phi(m,n)\\\partial^+_y\phi(m,n)\end{bmatrix}=
  \begin{bmatrix}\phi(m+1,n)-\phi(m,n)\\\phi(m,n+1)-\phi(m,n) \end{bmatrix}  \label{gradient},
\end{align}
where $m=1,\dots,M$ and $n=1,\dots,N$. For the Laplacian operator, we can use central difference to approximate it:
\begin{align}
    \Delta\phi(m,n)=&\partial_{x}^2\phi(m,n)+\partial_{y}^2\phi(m,n)\notag\\
    =&\phi(m-1,n)-2\phi(m,n)+\phi(m+1,n)\notag\\
    &+\phi(m,n-1)-2\phi(m,n)+\phi(m,n+1)\label{laplacian},
\end{align}
where $m=1,\dots,M$ and $n=1,\dots,N$. What's more, we also need to define $\divg(q)$. For any $q=\begin{bmatrix}q_1\\q_2\end{bmatrix}$, the divergence operator must satisfy
\begin{align}
\langle\nabla\phi,q\rangle=-\langle\phi,\divg(q)\rangle.
\end{align}
Therefore, the divergence should be approximated by
\begin{align}
    \text{div}(q)=\partial^-_x q_1+\partial^-_y q_2,
\end{align}
where 
\begin{align}
    \partial^-_x q_1(m,n)=\begin{cases}
    q_1(m,n)-q_1(m-1,n) &2\leq m\leq M\\
    q_1(1,n)-q_1(M,n) &m=1
    \end{cases}
\end{align}
for $n=1,2,\dots,N$, and
\begin{align}
    \partial^-_y q_2(m,n)=\begin{cases}
    q_2(m,n)-q_2(m,n-1) &2\leq n\leq N\\
    q_2(1,n)-q_2(m,N) &n=1
    \end{cases}
\end{align}
for $m=1,2,\dots,M$.

\section{Initialization for the algorithms}\label{appendix2}
In this section, we will briefly describe the ways to initialize $\phi$ in our proposed algorithms. One simple and efficient way is to approximate the original object by a polygon with few vertices. For the exact model, we rotate the given image by a certain degree for several times and then find the pixel at the top. We collect these pixels in a set $V_t$, and these points form a polygon $P_0$. We then initialize $\phi$ to be the SDF of $P_0$, which can be done by a very efficient MATLAB built-in function, called \textit{bwdist}. What's more, the points in $V_t$ can also serve as the boundary landmarks in (\ref{eq:convhull4}). For the Algorithm \ref{algo:outliers}, since the existence of outliers, we need to modify the initialization method a little. Instead of finding the top pixel each time, we find the pixel at the 5th or 10th percentile. These vertices can also form a good approximation to the object.

\bibliographystyle{siamplain}
\bibliography{references}
\end{document}